\documentclass[journal]{IEEEtran}
\usepackage{graphics} % for pdf, bitmapped graphics files
\usepackage{epsfig} % for postscript graphics files
\usepackage{setspace}
\usepackage{amssymb}
\usepackage{subfigure}
\usepackage{supertabular}
\usepackage{cite}
\usepackage{amsmath} %Kelli

\usepackage{float}
\usepackage{bm} %prev version
\usepackage{dblfloatfix}

\usepackage[]{graphicx}
\usepackage{amsmath} % assumes amsmath package installed

\usepackage{amssymb}  % assumes amsmath package installed
\usepackage{amsthm}
\usepackage{subfigure}
\usepackage{epsfig}
\usepackage{mathrsfs}
\usepackage{bm}
\usepackage{algpseudocode}
\usepackage{algorithm}
\usepackage{setspace}
\usepackage{graphics}
\usepackage{epsfig} % for postscript graphics files
\usepackage{multirow}
\usepackage{graphicx}
\usepackage[margin=1in]{geometry}
\usepackage{bm}
\usepackage{comment}
\usepackage{subfigure}
\usepackage{mathrsfs}
\title{A Gaussian Particle Filter Approach for Sensors to Track Multiple Moving Targets}

\author{H. Li \emph{} % <-this % stops a space
\thanks{The author H. Li is with Zhejiang University of Technology, Hangzhou, 310014, P.R.China . Email: zgdlhj@zjut.edu.cn.}%
}
\begin{document}

\maketitle
%%%%%%%%%%%%%%%%%%%%%%%%%%%%%%%%%%%%%%%%%%%%%%%%%%%%%%%%%%%%%
\begin{abstract}
In a variety of problems, the number and state of multiple moving targets are unknown and are subject to be inferred from their measurements obtained by a sensor
with limited sensing ability. This type of problems is raised in a variety of applications,
including monitoring of endangered species, cleaning, and
surveillance. Particle filters are widely used to estimate target state from its prior information and its measurements that recently become available, especially for the cases when the measurement model and the prior distribution of state of interest are non-Gaussian. However, the problem of estimating number of total targets and their state becomes intractable when the number of
total targets and the measurement-target association are unknown.
This paper presents a novel Gaussian particle filter technique that combines Kalman filter and particle filter
for estimating the number and state of total targets based on the measurement obtained online.
The estimation is represented by a set of weighted particles, different from classical particle filter, where each particle is a Gaussian distribution instead of a point mass.
\end{abstract}

%>>>> Include a list of keywords after the abstract

%%%%%%%%%%%%%%%%%%%%%%%%%%%%%%%%%%%%%%%%%%%%%%%%%%%%%%%%%%%%%
\section{INTRODUCTION}
\label{sec:introduction}
The problem of tracking and monitoring targets using
position-fixed sensors is relevant to a variety of applications,
including monitoring of moving targets using cameras
\cite{}, tracking anomalies in manufacturing
plants \cite{CullerOverviewSensorNetworks04}, and tracking of
endangered species \cite{LuRandomizedHybridSystemRRT10,JuanWildlifeTrackingZebraNet02,LuSPIE11}. The position-fixed
sensor is deployed to measure targets based on limited information that only becomes
available when the target enters the sensor's field-of-view (FOV) or
visibility region \cite{LuInformationPotential14}. The sensor's FOV is defined as a compact subset of
the region of interest, in which the sensor can obtain measurements
from the targets. In many such sensor applications, filter techniques are often required to estimate unknown variables of interest, for examples, target number and target state.

When the noise in the measurement model is an additive
Gaussian distribution, the target state can be estimated from
frequent observation sequence using a Kalman filter
\cite{WelchIntroductionKalmanFilter}. This approach is well suited to
long-range high-accuracy sensors, such as radars, and to moving
targets with a known dynamical model and initial conditions. However,
most of these underlying assumptions are violated in modern
applications, because the targets' motion models are unknown,
and, possibly, random and nonlinear. Also, due to the use of low-cost
passive sensors, measurement errors and noise may be non-additive and
non-Gaussian. An extended Kalman filter (EKF) can be used when the system dynamics
are nonlinear, but can be linearized about nominal operating
conditions \cite{JulierExtensionKalmanFilterNonlinearSystems97}.
An unscented Kalman filter (UKF) method, based on the unscented
transformation (UT) method, can be applied to compute the mean and
covariance of a function up to the second order of the Taylor
expansion
\cite{WanunscentedKalmanfilterNonlinearEstimation00}.

However, the efficiency of these filters decreases significantly when
the system dynamics are highly nonlinear or unknown, and when the measurement noise
are non-Gaussian. Recently, a non-parametric method based on
condensation and Monte Carlo simulation, known as a particle filter, has
been proposed for tracking multiple targets exhibiting nonlinear
dynamics and non-Gaussian random effects
\cite{ZiaMCMCBasedparticleFilterTrackingMultipleInteractingTargets04}.
Particle filters are well suited to modern surveillance systems
because they can be applied to Bayesian models in which the hidden
variables are connected by a Markov chain in discrete time. In the classical particle filter method, a weighted set of particles
or point masses are used to represent the probability density function (PDF) of the target state by
means of a superposition of weighted Dirac delta
functions. At each iteration
of the particle filter, particles representing possible target state
values are sampled from a proposal distribution \cite{ArulampalamTutorialParticleFilter02}. The weight
associated with each particle is then obtained from the target-state
likelihood function, and from the prior estimation of the target state
PDF. When the effective particle size is smaller than a predefined
threshold, a re-sampling technique can be implemented
\cite{CarpenterImprovedparticleFilterNonlinearProblems99}. One
disadvantage of classical particle-filtering techniques is that the
target-state transition function is used as the importance density
function to sample particles, without taking new observations into
account \cite{RuiBetterProposalDistributionsTrackingParticleFilter01}.
Recently, an particle filter with Mixture Gaussian representation was proposed by author for monitoring maneuvering targets \cite{luParticle14}, where the particles are sampled
based on the supporting intervals of the target-state likelihood
function and the prior estimation function of the target state. In this case,
the supporting interval of a distribution is defined as the $90\%$
confidence interval\cite{GormanTestSignificanceConfidenceInterval04}.
The weight for each particle is obtained by considering the likelihood
function and the transition function simultaneously. Then, the
weighted expectation maximization (EM) algorithm is implemented to use
the sampled weighted particles to generate a normal mixture model of
the distribution.

Kreucher proposed joint multitarget probability density (JMPD)\cite{KreucherMultitargetTrackingJMPD05} to estimate the number of total targets in workspace and their state, where targets are moving. By using JMPD, the data association problem is avoided, however, the JMPD results in a joint state space, the dimension of which is the dimension of a target state times number of total targets. Since the number of total targets is unknown, the joint space size remain unavailable. To overcome this problem, it is assumed the number of total targets has a maximum value. Therefore, when the maximum number of targets is large, the joint state space becomes intractable.

Inspired by \cite{WonKPF10}, this paper presents a novel filter technique which combines Kalman filter and particle filter
for estimating the number and state of total targets based on the measurement obtained online.
The estimation is represented by a set of weighted particles, different from classical particle filter, where each particle is a Gaussian instead of a point mass.
The weight of each particle represents the probability of existing a target, while its gaussian indicates the state distribution for this target.
More importantly, the update of particles is different from classical particle filter. For each particle, the gaussian parameters are updated based using Kalman filter given a measurement. To overcome the data association problem, in this paper, when one particle is updated, the other particles are considered as the measurement condition, which will be explained in Section {\ref{sec:method}}. The novel Gaussian particle filter technique requires less particles than classical particle filters, and can solve multiple target estimation problem without increasing the state space dimensions.

The paper is organized as follows. Section
\ref{sec:problemformulation} describes the multiple targets estimation problem formulation and assumptions.
The background on the
particle filter and Kalman filter is reviewed in Section
\ref{sec:background}. Section \ref{sec:method} presents the Gaussian Particle filter technique. The method is
demonstrated through numerical simulations and results, presented
in Section \ref{sec:results}. Conclusions and future work are described in
Section \ref{sec:conclusion}.

\section{Problem Formulation}
\label{sec:problemformulation}
There $N$  targets are moving in a two dimensional workspace denoted as $\mathcal{W}$ where $N$ denotes the unknown number of total targets. For simplicity, there is zero obstacle in the workspace. The goal of the sensor is to obtain the state estimation for all the targets, denoted as $\mathbf{X}_k$, and target number estimation, denoted as $T_k$, at time step $k$. The states for total targets at $k$ is denoted as $\mathbf{X}_k=[\mathbf{x}_k^1,\mathbf{x}_k^2,\cdots,\mathbf{x}_k^{N}]$ that has $N$ state vectors. The estimation of total target state at $k$ is denoted as $\mathbf{X}^k=[\mathbf{x}_k^1,\mathbf{x}_k^2,\cdots,\mathbf{x}_k^{T^k}]$. Let $T^k$ denote the estimation of total target number at $t^k$.
The $i$th target is modeled as
\begin{equation}
\label{eqn:targetModel}
\mathbf{x}_k^i=\mathbf{F}_k\mathbf{x}_{k-1}^i+\bm{\nu}_k,
\end{equation}
where
\begin{equation}
\bm{\nu}_k\sim N(0,\mathbf{Q}_k).
\end{equation}
Furthermore, $\mathbf{F}_k$ and $\mathbf{Q}_k$ are assumed known.

In standard estimation theory, a sensor that obtains a vector of
measurements $\mathbf{z}^k \in \mathbb{R}^r$ in
order to estimate an unknown state vector set $\mathbf{X}^k \in \mathbb{R}^n$ at time $k$ is modeled as,
\begin{equation}
\label{eqn_SensorModelEstimation}
\mathbf{z}^k = \mathbf{h}(\mathbf{X}^k,\bm{\lambda}^k),
\end{equation}
where $\mathbf{h}: \mathbb{R}^{n+\wp} \rightarrow \mathbb{R}^r$ is a
deterministic vector function that is possibly nonlinear, the random
vector $\bm{\lambda}^k \in \mathbb{R}^{\wp}$ represents the sensor
characteristics, such as sensor action, mode \cite{LuADPcdc13}, environmental conditions, and
sensor noise or measurement errors. In this paper, the sensor is modeled as
\begin{equation}
\label{eqn:SensorModel}
\mathbf{z}_k=\frac{1}{N}\sum_{i=1}^{N}\mathbf{x}_i+\bm{\omega}_k.
\end{equation}
It is further assumed that the whole workspace is visible to a position fixed sensor (not shown).

\section{Background}
\label{sec:background}
\subsection{Particle Filter Methods}
The particle filter is a recursive model estimation method based on sequential Monte Carlo Simulations. Because of their recursive nature, particle filters are easily
applicable to online data processing and variable inference. More importantly, it is applicable to
nonlinear system dynamics with non-Gaussian noises. The PDF functions
are represented with properly weighted and relocated point-mass, known as particles.
These particles are sampled from an importance density that is
crucial to the particle filter algorithm and is also referred to as a proposal distribution. Let
$\{\mathbf{x}^{\kappa}_{j,p},w^{\kappa}_{j,p}\}^{N}_{p=1}$ denote the weighted
particles that are used to approximate the posterior PDF $f(\mathbf{x}^{\kappa}_j~|~Z^{\kappa}_j)$ for the $j$th
target at $t_{\kappa}$, where $Z^{\kappa}_j = \{\mathbf{z}^{0}_j, \ldots,
\mathbf{z}^{\kappa}_j\}$ denotes the set of all measurements obtained by
sensor $i$, from target $j$, up to $t_{\kappa}$. Then, the posterior probability density function of
the target state, given the measurement at $t_{\kappa}$ can be modeled as,
\begin{equation} f(\mathbf{x}^{\kappa}_j~|~Z^{\kappa}_j)=\sum^N_{p=1}
w^{\kappa}_{j,p}\delta(\mathbf{x}^{\kappa}_{j,p}), ~~\sum^N_{p=1}w^{\kappa}_{j,p}=1
\end{equation}
where $w^{\kappa}_{j,p}$ is non-negative and $\delta$ is the Dirac delta
function. The techniques always
consist of the recursive propagation of the particles and the particle
weights. In each iteration, the particles $\mathbf{x}^{\kappa}_{j,p}$ are
sampled from the importance density $q(\mathbf{x})$. Then, weight
$w^k_{j,p}$ is updated for each particle by
\begin{equation}
w^{\kappa}_{j,p}\propto \frac{p(\mathbf{x}^{\kappa}_{j,p})}{q(\mathbf{x}^{\kappa}_{j,p})}
\end{equation}
where $p(\mathbf{x}^{\kappa}_{j,p})\propto
f(\mathbf{x}^{\kappa}_{j,p}~|~Z^{\kappa}_j)$. Additionally, the weights are
normalized at the end of each iteration.

One common drawback of particle filters is the
degeneracy
phenomenon\cite{RuiBetterProposalDistributionsTrackingParticleFilter01},
i.e., the variance of particle weights accumulates along iterations.
A common way to evaluate the degeneracy phenomenon is the
effective sample size
$N_{e}$\cite{CarpenterImprovedparticleFilterNonlinearProblems99},
obtained by,
\begin{equation}
N_e=\frac{1}{\sum^N_{p=1}(w^{\kappa}_{j,p})^2}
\end{equation}
where $w^{\kappa}_{j,p}, ~p=1,2,\dots,N$ are the normalized weights. In general, a re-sampling
procedure is taken when $N_e <N_s$, where $N_s$ is a predefined
threshold, and is usually set as $\frac{N}{2}$. Let
$\{\mathbf{x}^{\kappa}_{j,p},w^{\kappa}_{j,p}\}^{N}_{p=1}$ denote the particle set that
needs to be re-sampled, and let
$\{\mathbf{x}^{{\kappa}*}_{j,p},w^{{\kappa}*}_{j,p}\}^{N}_{p=1}$ denote the particle set
after re-sampling. The main idea of this re-sampling procedure is to
eliminate the particles having low weights by re-sampling
$\{\mathbf{x}^{{\kappa}*}_{j,p},w^{{\kappa}*}_{j,p}\}^{N}_{p=1}$ from
$\{\mathbf{x}^{\kappa}_{j,p},w^{\kappa}_{j,p}\}^{N}_{p=1}$ with the probability of
$p(\mathbf{x}^{{\kappa}*}_{j,p}=\mathbf{x}^{\kappa}_{j,s})=w^{\kappa}_{j,s}$. At the end of the resampling
procedure, $w^{{\kappa}*}_{j,p}, p=1,2,\dots,N$ are set as $1/N$.
\subsection{Kalman Filter Methods}
The well known Kalman filter is also a recursive method to estimate system/target state based on a measurement sequence, minimizing the estimation uncertainty. The measurement of the system state with an additive Gaussian noise are given by the sensor. Then, in each iteration, the Kalman filter consists of two precesses: i) it predicts the system state and their uncertainties; ii) it updates the system state and uncertainties with the measurement that newly becomes available.
The system dynamics is given  as
\begin{equation}
\mathbf{x}_k=\mathbf{F}_k\mathbf{x}_{k-1}+\mathbf{B}_k\mathbf{u}_k+\bm{\nu}_k
\end{equation}
where subscript $k$ and $k-1$ denote the current and previous time index, while $\mathbf{F}_k$ is the system discrete transition matrix, and  $\mathbf{B}_k$ and $\mathbf{u}_k$ are the control matrix and control input. $\bm{\nu}_k$ is the white noise, defined as
\begin{equation}
\bm{\nu}_k\approx N(0,\mathbf{Q}_k)
\end{equation}
where $\Sigma_k$ is the covariance.
At $k$th time step, an measurement of the system true state $\mathbf{x}_k$ is made by a sensor, is given by
\begin{equation}
\label{eqn:KalmanZ}
\mathbf{z}_k=\mathbf{H}_k\mathbf{x}_k+\bm{\omega}_k
\end{equation}
where $\mathbf{H}_k$ is a mapping from system state space to measurement space, and white noise $\mathbf{Q}_k$ is defined as
\begin{equation}
\bm{\omega}_k\approx N(0,R_k)
\end{equation}
It is assumed that the noise $\bm{\omega}_k$ and $\bm{\nu}$ at each time step are independent.

Let $\tilde{\mathbf{x}}_k$ denote the predicted state estimation given $\hat{\mathbf{x}}_{k-1}$, where $\hat{\mathbf{x}}_{k-1}$ is the updated estimation of system state at $k-1$ time step.
Furthermore, let $\tilde{\bm{\Sigma}}_k$ denote the predicted covariance given $\hat{\bm{\Sigma}}_{k-1}$, where $\hat{\bm{\Sigma}}_{k-1}$ is the updated estimation covariance.
Then, in the predicting step,
\begin{align}
\tilde{\mathbf{x}}_k=\mathbf{F}_k\hat{\mathbf{x}}_{k-1}+\mathbf{B}_k\mathbf{u}_k\\
\tilde{\bm{\Sigma}}_k=\mathbf{F}_k\hat{\bm{\Sigma}}_{k-1}\mathbf{F}^T_k+\mathbf{Q}_k
\end{align}
In the updating step, the measurement $\mathbf{z}_k$ is used, together with above predicted state and covariance, to update the state and covariance.
The residual, $\mathbf{y}_k$ between measurement and predicted state is given by
\begin{equation}
\mathbf{y}_k=\mathbf{z}_k-\mathbf{H}_k\tilde{\mathbf{x}}_k
\end{equation}
The innovation covariance $\mathbf{S}_k$ is given by
\begin{equation}
\mathbf{S}_k=\mathbf{H}_k\tilde{\bm{\Sigma}}_k\mathbf{H}^T_k+\mathbf{R}_k
\end{equation}
Then, the optimal Kalman gain is calculated as
\begin{equation}
\mathbf{K}_k=\tilde{\bm{\Sigma}}_k\mathbf{H}^T_k\mathbf{S}^{-1}_k
\end{equation}
Then, the state and covariance can be updated by
\begin{align}
\hat{\mathbf{x}}_{k}=\tilde{\mathbf{x}}_k+\mathbf{K}_k\mathbf{y}_k\\
\hat{\bm{\Sigma}}_{k}=(\mathbf{I}-\mathbf{K}_k\mathbf{H}_k)\tilde{\bm{\Sigma}}_k
\end{align}

%%%%%%%%%%%%%%%%%%%%%%%%%%%%%%%%%%%%%%%%%%%%%%%%%%%%%%%%%%%%%

\section{Methodology}
\label{sec:method}
In this paper, a novel Gaussian particle filter technique follows the main idea of particle filter
for estimating the number and state of total targets based on the measurement obtained online.
Different from classical particle filter, each particle here is a gaussian instead of a point mass.
The estimation for number of total targets and their state is presented by a set of weighted particles. The $i$th particle at time $k$ is denoted as
\begin{equation}
P^i_k=\{w^i_k,\mathcal{N}(\mathbf{x}^i_k|\bm{\mu}^i_k,\bm{\Sigma}^i_k)\}
\end{equation}
where $w_i$ is the probability of existing a target having a state distribution as $N(\bm{\mu},\mathcal{N}(\mathbf{x}_i|\bm{\mu},\bm{\Sigma}_i))$.
By this particle definition, the dimensions of the system state remains the same as the dimensions of each individual target.
When these particles are available, the estimated number of total targets can be given as
$T=\sum_{i=1}^{N_p}w_i$,
where $N_p$ is the particle number.

Notice that the particle representation is different from classical particle filter, where each particle represents a possible value of system state. The updating of each particle and total weights are also different from classical particle filter. Kalman filter is used to update each particle, the weight and the distribution. Notice that since the measurement at each time step is conditioned on all the targets in the FOV, while in the classical Kalman filter method one measurement is associated with one target, which means data association problem is avoid. Therefore, the Kalman filter is modified to updated the particles which are coupled by one measurement, and some approximations and assumptions are further needed.

Similar to Kalman filters and particle filters, the algorithm proposed in this paper is a recursive method. Assume that at time step $k$, the measurement $\bm{z}_k$ is available, and the estimation of the system at time step $k-1$ is represented by a particle set, denoted as $\mathcal{P}_{k-1}=\{P^1_{k-1},P^2_{k-1},\dots,P^{N_p}_{k-1}\}$, where $N_p$ is the number of all particles. By using the target dynamic function \ref{eqn:targetModel}, $\mathcal{P}_{k-1}$ can be updated to $\tilde{\mathcal{P}}_{k}$ without using the $\bm{z}_k$. Due to limit of FOV, only a few particles may have contribution to the measurement. Let $\mathcal{P}_S$ denote set including the particles lie in the FOV, while let $\bar{\mathcal{P}}=\tilde{\mathcal{P}}_{k}/\mathcal{P}_S$ denote the compensation set. Only the particles in $\mathcal{P}_S$ are updated. Please Note the size of $\mathcal{P}_S$ is small.
Without generality, assume that $\mathcal{P}_S=\{\tilde{P}^1_k,\tilde{P}^2_k,\dots,\tilde{P}^s_k\}$, where $s$ the number of particles
The update of each particle in $\mathcal{P}_S$ is calculated separately.
Without generally, we focus on updating $\tilde{P}^j_k$,
Let a boolean set $E=[e_1,e_1,\dots,e_s]$, where $e_i\in\{0,1\}$. For any $E$ with $e_j=1$ such that $\Pi (w_i)^{e_i}(1-w_i)^{1-e_i}>\epsilon$, where $\epsilon$ is a predefined threshold, a particle is calculated and denoted as $\tilde{p}_j$,
Then, the modified Kalman filter is used to give the updated gaussian parameters of all particles with $e_i=1$.
According to sensor model \ref{eqn:SensorModel}, the measurement is given by
\begin{equation}
\bm{z}_k=\frac{\sum_{i=1}^{s}\bm{\mu}^i_k e_i}{\sum_{i=1}^{s}{e_i}}+\frac{1}{(\sum_{i=1}^{s}{e_i})^2}\sum_{i=1,i\ne j}^{s}{e_i}\bm{\Sigma_i}+\bm{\omega}_k
\end{equation}
Compare the above function to (\ref{eqn:KalmanZ}), we have following setting
\begin{eqnarray}
\bm{H}_k=I\times\frac{1}{(\sum_{i=1}^{s}{e_i})}\\
\bm{z}_k=\bm{z}_k-\frac{\sum_{i=1,i\ne j}^{s}\bm{\mu}^i_k e_i}{\sum_{i=1}^{s}{e_i}}\\
\bm{R}_k=\frac{1}{(\sum_{i=1}^{s}{e_i})^2}\sum_{i=1,i\ne j}^{s}{e_i}\bm{\Sigma_i}+\bm{\omega}_k
\end{eqnarray}
Then, by applying Kalman procedure
\begin{eqnarray}
\mathbf{y}_k=\mathbf{z}_k-\mathbf{H}_k\bm{\mu}_k\\
\mathbf{S}_k=\mathbf{H}_k\bm{\Sigma}_k\mathbf{H}^T_k+\mathbf{R}_k\\
\mathbf{K}_k=\bm{\Sigma}_k\mathbf{H}^T_k\mathbf{S}^{-1}_k\\
\bm{\mu}_{k}=\bm{\mu}_k+\mathbf{K}_k\mathbf{y}_k\\
\bm{\Sigma}_{k}=(\mathbf{I}-\mathbf{K}_k\mathbf{H}_k)\bm{\Sigma}_k
\end{eqnarray}
Its proof can be found in the appendix.

Once each particle appearing in combination $E$ has been updated,
the weight $w_c$ is for the particle combination $E$ can be obtained by
\begin{eqnarray}
w_c&=&\Pi (w_i)^{e_i}(1-w_i)^{1-e_i}
\times\frac{1}{(2\pi)^2\|\bm{\Sigma}^{-1}_c\|}\nonumber\\&&\times\exp\{-(\mathbf{z}_k-\bm{\mu}_c)^T\bm{\Sigma}^{-1}_c(\mathbf{z}_k-\bm{\mu}_c)\}
\end{eqnarray}
where $c\in I_E$, where $I_E$ is the combination index, and $\bm{\mu}_c)$ and $\bm{\Sigma}^{-1}_c$ is given by
\begin{eqnarray}
\bm{\mu}_c=\mathbf{H}_k\sum\bm{\mu}^i_{k}\\
\bm{\Sigma}_c=\mathbf{H}_k\sum\bm{\Sigma}^i_{k}\mathbf{H}^T_k+\mathbf{R}_k
\end{eqnarray}
Then, insert particle $\{\mathcal{N}^i(\bm{\mu}_{k},\bm{\Sigma}_{k})\}$ into a set $G_c$ for combination $c\in I_E$, the set $G_c$ has a weight $w_c$.
After all $G_c$ the combination of $E$ that $\Pi (w_i)^{e_i}(1-w_i)^{1-e_i}>\epsilon$ is obtained.
Weights are updated by
\begin{equation}
w_c=\frac{w_c}{\sum w_c}
\end{equation}.
Then in each group $G_c$, the weight of $i$th particle is updated as
\begin{equation}
w^i_c=\frac{w^i_k}{\Pi w^i_k}*w_c
\end{equation}

The particles in all set $G_c$ are updated from the same particle in the previous set.
If two particles in is close enough, then they are combined as one particle, the weight of which is set as the summation of both weights.
The distance between $\bm{\mu}_i$ and $\bm{\mu}_j$ is defined as Mahal-distance
\begin{equation}
(\bm{\mu}_i-\bm{\mu}_j)^T\bm{M}(\bm{\mu}_i-\bm{\mu}_j)
\end{equation}
and its covariance is updated as
\begin{equation}
\bm{\Sigma}=\sum\bm{\Sigma}^i_{k}
\end{equation}

\section{Simulation and Results}
\label{sec:results}
\begin{figure}[h]
\centering
    \includegraphics[width=3in]{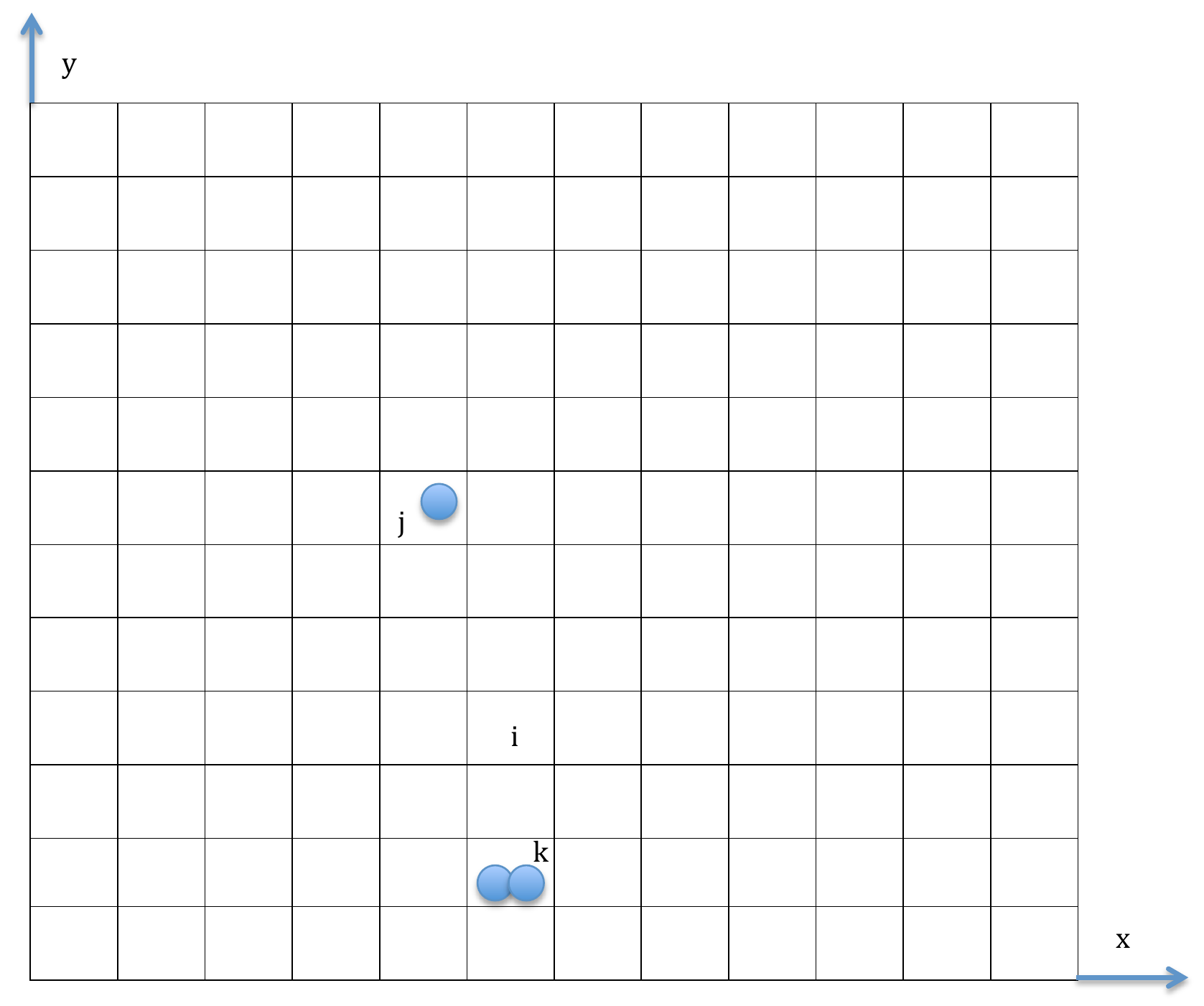}
\begin{minipage}{0.5\textwidth}
 \caption{The workspace contians three point targets.}
 \label{fig:workspace_cell}
\end{minipage}
\end{figure}

As shown in figure (\ref{fig:workspace_cell}), $N$ targets, represented by blue dots, are moving in the $2$ dimensional workspace. The whole workspace is visible to a position fixed sensor (not shown) and the workspace is discretized into $12\times12$ cells. Each cell represents a $1\times1$ rectangular area. The $i$th cell, denoted as $C_i, i\in\mathcal{C}$, is defined by $[x^{ul}_i, y^{ul}_i, x^{dr}_i, y^{dr}_i ]$, where $\mathcal{C}$ is the cell index set and $(x^{ul}_i, y^{ul}_i)$ and $(x^{dr}_i, y^{dr}_i)$ are up left and down right corner coordinates of the $i$th rectangular area respectively. Only $M$ cells can be measured at each time step $k$, and they don't have to be adjacent. The goal of the sensor is to estimate the target states and target number at time $k$. In this paper, information value function based $\alpha$ divergence is used to select the best $M$ cells to measure at each step \cite{LuInfoMove12}. The estimation of target states and target number at time $k$ is represented by joint multitarget probability density(JMPD) and it is updated after obtaining new measurements.

The target time-discrete state transition function can be written as
\begin{equation} % requires amsmath; align* for no eq. number
 \label{eqn:TargetDiscreteTransition}
 \mathbf{x}^{k+1}_i=\mathbf{F}\mathbf{x}^{k}_i+\mathbf{w}^k_i
\end{equation}
where
\begin{equation}
   \mathbf{F}=\left[
   \begin{matrix} % or pmatrix or bmatrix or Bmatrix or ...
      1 & \tau& 0 &0\\
      0 & 1&0 &0 \\
      0& 0& 1&\tau \\
      0&0&0&1
   \end{matrix}\right]
\end{equation}
and $\mathbf{w}^k_i$ is $0$ mean Gaussian noise with covariance $\mathbf{Q}=$diag$(20,0.2,20,0.2)$, and $\tau$ is the time step length, and $i\in \{1,2,\cdots, T^k\}$\cite{KreucherMultitargetTrackingJMPD05}.

It is further assumed that i) the sensor can measure any cell at time $k$; ii) the sensor can only measure up to $M$ cells at time $k$. The sensor condition
$\bm{\lambda}^k_c$ represents the signal to  noise ratio $\mbox{SNR}$, currently, it has only one possible value, fixed and known.
The measurement $z^k_i$ is a discrete variable, then joint PMF can be written as
\begin{equation}
\label{eqn:factorizationP}
f(\mathbf{z}^k,\mathbf{X}^k,T^k, \bm{\lambda}^k) \!=\!
f(\mathbf{z}^k|\mathbf{X}^k, T^k,\bm{\lambda}^k) f(\mathbf{X}^k, T^k)
f(\bm{\lambda}^k)
\end{equation}

 When measuring a cell, the imager sensor will give a Raleigh return, either a $0$ (no detection) or a $1$ (detection) governed by detecting probability, denoted as $p_d$, and false alarm probability, denoted as $p_f$. According to standard model for threshold detection of Rayleigh returns, $p_f=p_d^{(1+\mbox{\footnotesize SNR})}$. When $T$ targets are in the same cell, then the detection probability is $p_d(T)=p_d^{(1+\mbox{\footnotesize SNR})/(1+T\times \mbox{\footnotesize SNR})}$ and the $i$th sensor measurement at time $k$ can be evaluated by
\begin{eqnarray}
 \nonumber &&p(z^k_i|\mathbf{X}^k,T^k, \lambda^k_{a, i},\bm{\lambda}^k_c)=\begin{cases} p_d(T) & z^k_i=1 \\1-p_d(T) & z^k_i=0 \end{cases}
 \\&& \nonumber T=\sum^N_{j=1} (x^k_j\geq x^{ul}_c)\cap(x^k_j<x^{dr}_c)\nonumber\\
 &&\quad\quad\cap(y^k_j\geq y^{ul}_c)\cap(x^k_j<x^{dr}_c), ~c=\lambda^k_{a, i}
 \\&& p_d(T)=p_d^{(1+\bm{\lambda}^k_c)/(1+T\bm{\lambda}^k_c)}
 \end{eqnarray}
where $x^k_j, y^k_j$ are two position components of $\mathbf{x}^k_j\in\mathbf{X}^k$ and $T^k$ is the target number. Additionally, operators "$\geq$" and "$<$" return either $1$ if true or $0$ if false, while "$\cap$" is the Boolean  operator "and". For example, as shown in figure [\ref{fig:workspace_cell}], when $c=k$, $T=2$, similarly, when $c=j(i)$, $T=1(0)$.

\begin{figure}[h]
\centering
    \includegraphics[width=3in]{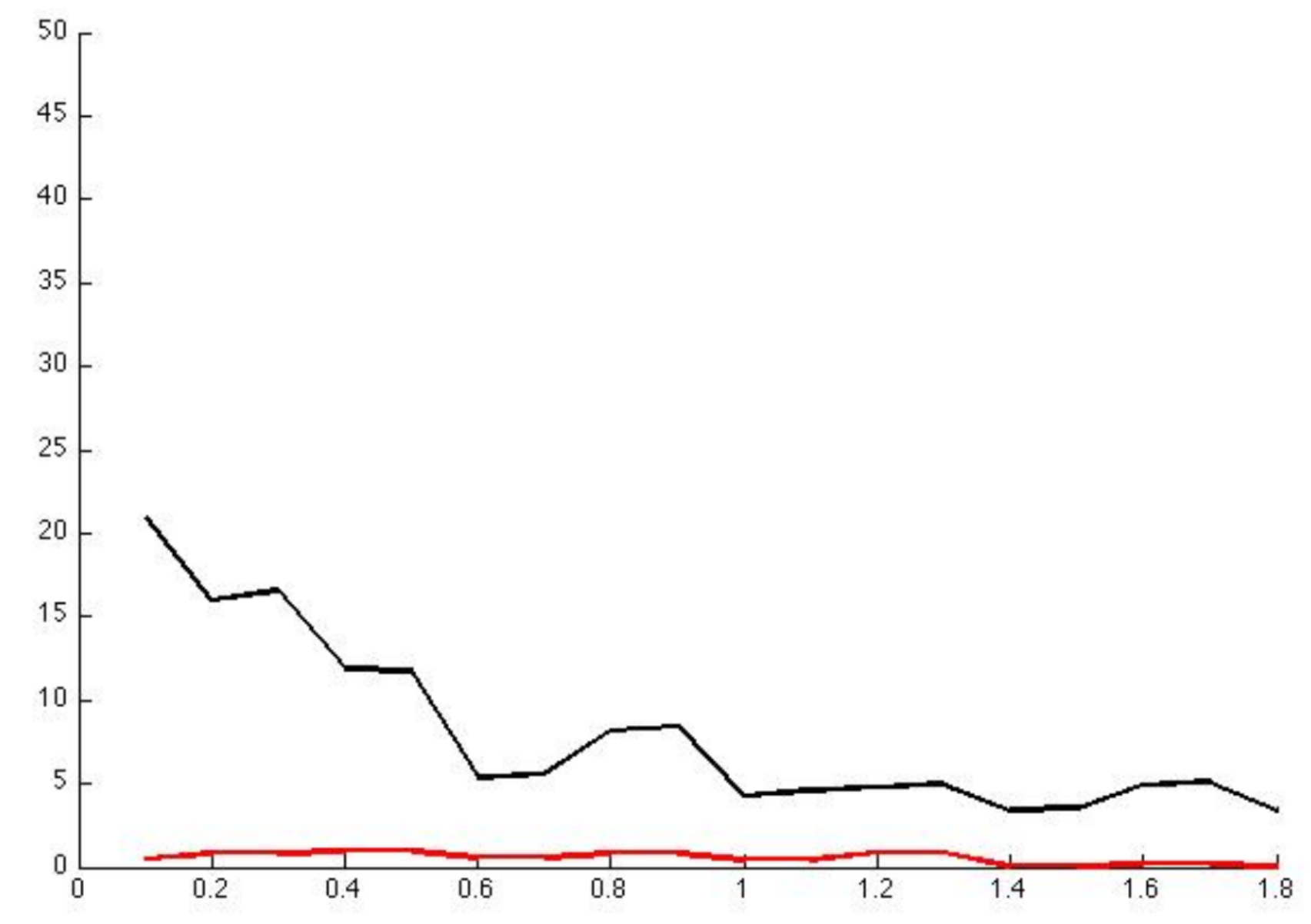}
 \caption{Simulation Result}
 \label{fig:result}
\end{figure}

\begin{figure*}[h]
\centering
    \includegraphics[width=5in]{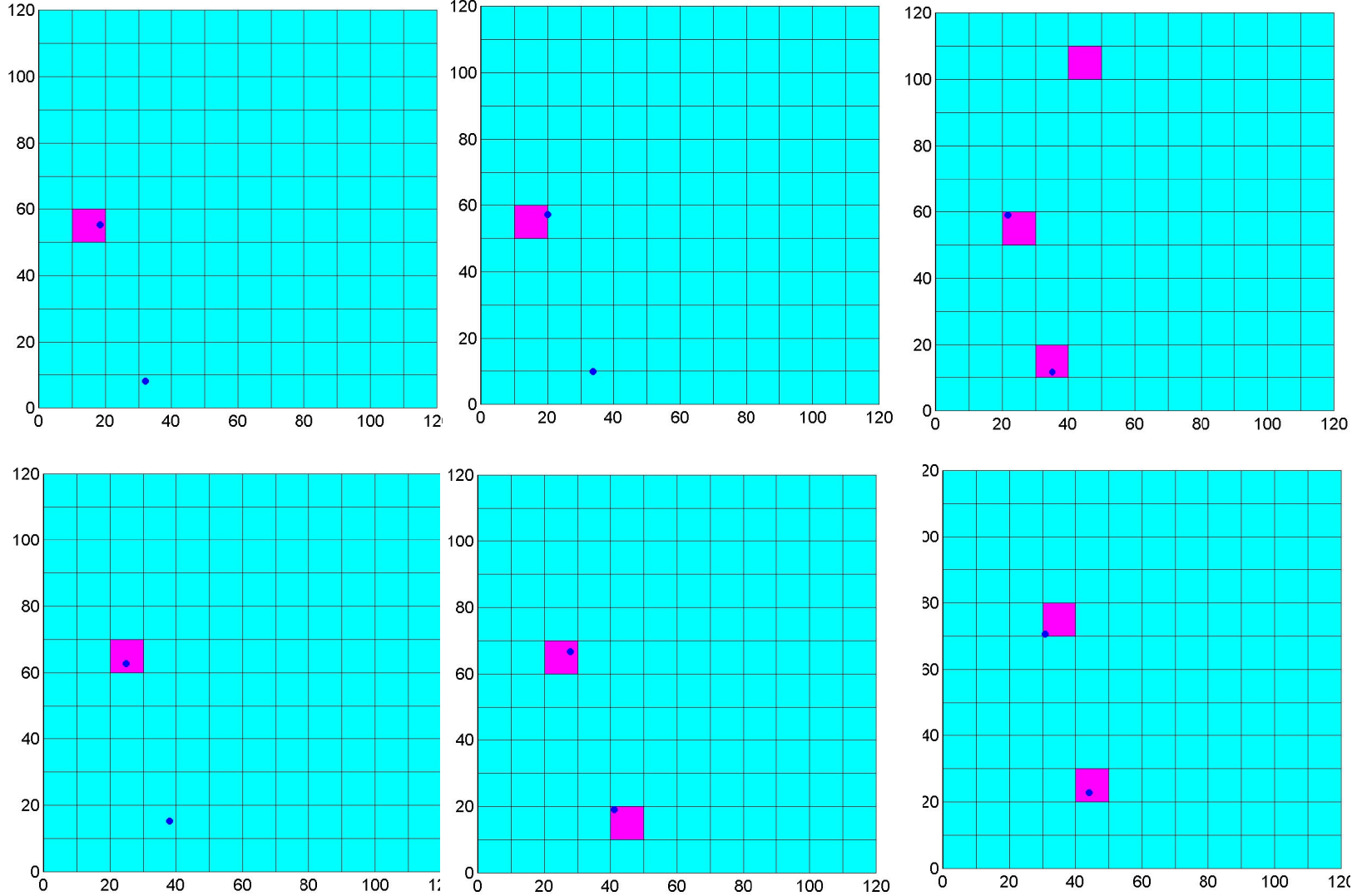}
 \caption{Snapshot of simulations}
 \label{fig:figsnap}
\end{figure*}

A snapshot of simulations is shown in Fig. \ref{fig:figsnap}, where magenta squares represent positive measurement return and blue dots represent the true targets' positions. The simulation results are summarized in Fig.  \ref{fig:result}, where the black curve represents the target state estimation error and the red curve represents the target number estimation error. As shown in Fig. \ref{fig:result}, both errors decreases as more measurements become available.

\section{Conclusion and Future Work}
\label{sec:conclusion}
A Gaussian particle filter that combines Kalman filter and particle filter is presented in this paper
for estimating the number and state of total targets based on the measurement obtained online.
The estimation is represented by a set of weighted particles, different from classical particle filter, where each particle is a Gaussian instead of a point mass.
The weight of each particle represents the probability of existing a target, while its Gaussian indicates the state distribution for this target. This approach is efficient for the problem of estimating number of total targets and their state.
\section{Appendix}
Without losing generality, $\mathcal{P}_S=\{\tilde{P}^1_k,\tilde{P}^2_k,\dots,\tilde{P}^s_k\}$, $E=[e_1,e_2,\dots,e_s]$,
for any particle such that $e_j=1$, its $\bm{\mu}^j_k$ and $\bm{\Sigma}^j_k$, given $\bm{z}_k$ and $\mathcal{P}_S$.
\begin{equation}
y_k=\bm{z}_k-\frac{\sum_{i=1}^{s}\bm{x}^i_k e_i}{\sum_{i=1}^{s}{e_i}})
\end{equation}
\begin{eqnarray}
\bm{\Sigma^j_k}&=&\mbox{COV}(\bm{x}^j_k-\hat{\bm{x}}^j_k)\nonumber\\
&=&\mbox{COV}(\bm{x}^j_k-(\tilde{\bm{x}}^j_k+\bm{K}^j_k y_k))\nonumber\\
&=&\mbox{COV}\Big(\bm{x}^j_k-(\tilde{\bm{x}}^j_k+\bm{K}^j_k (\frac{\sum_{i=1}^{s}\bm{x}^i_k e_i}{\sum_{i=1}^{s}{e_i}})\nonumber\\
&&+\bm{\nu}_k-\frac{\sum_{i=1}^{s}\tilde{\bm{x}}^i_k e_i}{\sum_{i=1}^{s}{e_i}})\Big)\nonumber\\
&=&\mbox{COV}\Big((\bm{I}-\frac{1}{\sum_{i=1}^{s}{e_i}}\bm{I}\bm{K}^j_k)(\bm{x}^j_k-\tilde{\bm{x}}^j_k)\nonumber\\
&&-\bm{K}^j_k\bm{\nu}_k-\sum_{i=1,i\ne j}^{s}\bm{K}^j_k(\bm{x}^i_k-\tilde{\bm{x}}^i_k)\Big)\\
&=&(\bm{I}-\frac{1}{\sum_{i=1}^{s}{e_i}}\bm{I}\bm{K}^j_k)\tilde{\Sigma}^j_k(\bm{I}-\frac{1}{\sum_{i=1}^{s}{e_i}}\bm{I}\bm{K}^j_k)^T
\nonumber\\
&&+\frac{1}{\sum_{i=1}^{s}{e_i}}\bm{I}\bm{K}^j_k\sum_{i=1,i\ne j}^{s}\bm{\Sigma}^i_k(\frac{1}{\sum_{i=1}^{s}{e_i}}\bm{I}\bm{K}^j_k)^T\nonumber\\
&&+\bm{K}^j_k\bm{R}_k\bm{K}^j_k.
\end{eqnarray}
By setting
$\partial_{\partial \bm{K}^j_k}=0$,
therefore
\begin{eqnarray}
\bm{K}^j_k&=&\bm{\Sigma}^j_k(\frac{1}{\sum_{i=1}^{s}{e_i}}\bm{I})^T\nonumber\\
&&\times(\bm{R}_k+\frac{1}{\sum_{i=1}^{s}{e_i}}\bm{I}\sum_{i=1}^{s}\bm{\Sigma}^i_k(\frac{1}{\sum_{i=1}^{s}{e_i}}\bm{I})^T)^{-1}\nonumber\\
&=&\frac{1}{\sum_{i=1}^{s}{e_i}}\!\bm{\Sigma}^j_k(\bm{R}_k\!+\!\frac{1}{(\sum_{i=1}^{s}{e_i})^2}\sum_{i=1}^{s}\bm{\Sigma}^i_k)^{-1}.
\end{eqnarray}
Then,
\begin{equation}
\bm{\mu}^j_k=\tilde{\bm{\mu}}^j_k+\bm{K}^j_ky_k
\end{equation}
\begin{eqnarray}
\bm{\Sigma}^j_k&=&\tilde{\bm{\Sigma}}^j_k-\frac{1}{\sum_{i=1}^{s}{e_i}}\tilde{\bm{\Sigma}}^j_k\nonumber\\
&&\times(\bm{R}_k\!+\!\frac{1}{\sum_{i=1}^{s}{e_i}}\bm{I}\sum_{i=1}^{s}\bm{\Sigma}^i_k(\frac{1}{\sum_{i=1}^{s}{e_i}}\bm{I})^T)^{-1}\nonumber\\
&&\times\tilde{\bm{\Sigma}}^j_k
\end{eqnarray}
\bibliography{WenjieLuPaper,ADP,wenjiebib1,wenjiebib,siliva_refs,rafael_refs_all}   %>>>> bibliography data in report.bib
\bibliographystyle{IEEEtran}   %>>>> makes bibtex use spiebib.bst

\end{document}